\documentclass{article}

\usepackage{microtype}
\usepackage{graphicx}
\usepackage{subcaption}
\usepackage{booktabs} 
\usepackage{multirow}
\usepackage[table]{xcolor}

\usepackage{hyperref}

\usepackage[preprint]{icml2026}

\usepackage{amsmath}
\usepackage{amssymb}
\usepackage{mathtools}
\usepackage{amsthm}

\usepackage[capitalize,noabbrev]{cleveref}

\theoremstyle{plain}

\theoremstyle{definition}

\theoremstyle{remark}

\usepackage[textsize=tiny]{todonotes}

\newcommand{\model}{\textsc{AGFF-Embed}}

\icmltitlerunning{Adaptive Global and Fine-Grained Perceptual Fusion for MLLM Embeddings Compatible with Hard Negative Amplification}

\begin{document}

\twocolumn[
  \icmltitle{Adaptive Global and Fine-Grained Perceptual Fusion for MLLM Embeddings Compatible with Hard Negative Amplification}



  \icmlsetsymbol{equal}{*}
  \icmlsetsymbol{x}{\dag}
  \icmlsetsymbol{y}{\ddag}

  \begin{icmlauthorlist}
    \icmlauthor{Lexiang Hu}{a,x}
    \icmlauthor{Youze Xue}{b}
    \icmlauthor{Dian Li}{b,y}
    \icmlauthor{Gang Liu}{b}
    \icmlauthor{Zhouchen Lin}{a,c}
  \end{icmlauthorlist}

  \icmlaffiliation{a}{State Key Lab of General AI, School of Intelligence Science and Technology, Peking University}
  \icmlaffiliation{b}{Tecent QQ}
  \icmlaffiliation{c}{Institute for Artificial Intelligence, Peking University}
  
  \icmlcorrespondingauthor{Dian Li}{goodli@tecent.com}
  \icmlcorrespondingauthor{Zhouchen Lin}{zlin@pku.edu.cn}

  \icmlkeywords{MLLM Embeddings}

  \vskip 0.3in
]



\printAffiliationsAndNotice{
    \textsuperscript{$\dag$}Work done during internship at Tencent QQ \textsuperscript{$\ddag$}Project leader }  

\begin{abstract}
  Multimodal embeddings serve as a bridge for aligning vision and language, with the two primary implementations---CLIP-based and MLLM-based embedding models---both limited to capturing only global semantic information. Although numerous studies have focused on fine-grained understanding, we observe that complex scenarios currently targeted by MLLM embeddings often involve a hybrid perceptual pattern of both global and fine-grained elements, thus necessitating a compatible fusion mechanism. In this paper, we propose \textbf{A}daptive \textbf{G}lobal and \textbf{F}ine-grained perceptual \textbf{F}usion for MLLM \textbf{Embed}dings (\model), a method that prompts the MLLM to generate multiple embeddings focusing on different dimensions of semantic information, which are then adaptively and smoothly aggregated. Furthermore, we adapt \model~with the Explicit Gradient Amplification (EGA) technique to achieve in-batch hard negatives enhancement without requiring fine-grained editing of the dataset. Evaluation on the MMEB and MMVP-VLM benchmarks shows that \model~comprehensively achieves state-of-the-art performance in both general and fine-grained understanding compared to other multimodal embedding models.
\end{abstract}

\section{Introduction}

\begin{figure*}[t]
  \begin{center}
    \centerline{\includegraphics[width=\textwidth]{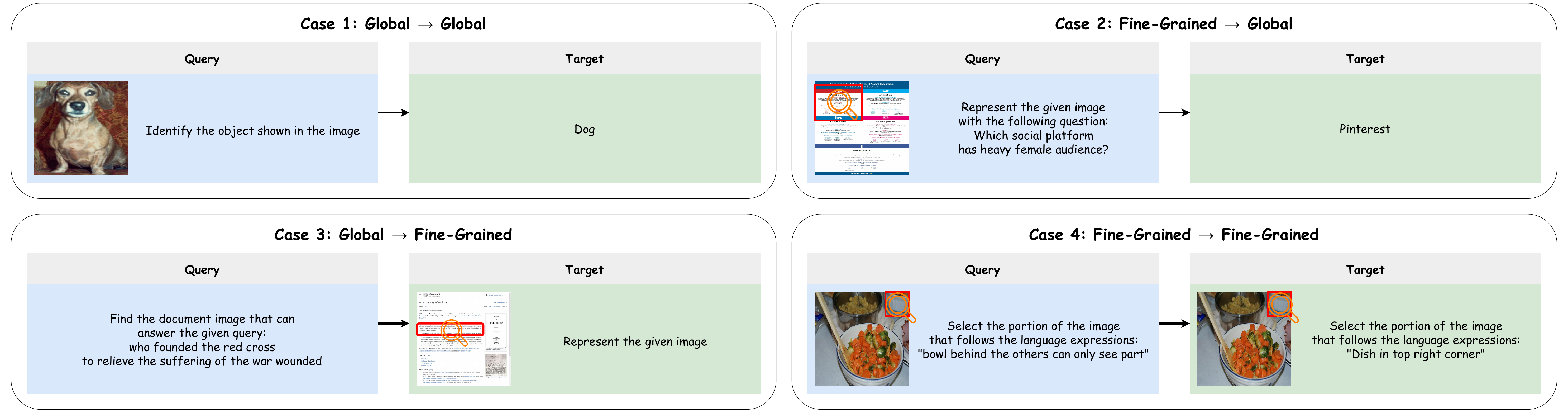}}
    \caption{
      For image-text-to-image-text matching tasks, whether the query and target should focus on global or fine-grained semantic information varies case by case.
    }
    \label{fig:global-or-fine-grained}
  \end{center}
  \vskip -0.2in
\end{figure*}

The learning of embedding space has always been a critical issue in the multimodal field. Early CLIP \cite{radford2021learning} utilizes contrastive learning with massive amounts of image-caption pairs to map images and text into the same semantic space, thereby achieving strong zero-shot transfer capabilities. Subsequent works have proposed variants of CLIP by improving data engineering \cite{xu2024demystifying}, scaling model parameters \cite{sun2023eva,sun2024eva}, and optimizing loss functions \cite{zhai2023sigmoid,tschannen2025siglip}, effectively promoting its maturity. However, CLIP-based models are limited to the simple form of single image-single caption pairs and fail to fully leverage the increasingly powerful capabilities of pretrained multimodal large language models (MLLMs) \cite{liu2023visual,li2024llava,an2025llava,zhu2025internvl3}. Recently, VLM2Vec \cite{jiang2025vlm2vec} extracts embeddings from image-text mixed content based on state-of-the-art MLLMs, achieving outstanding performance in a range of complex image-text-to-image-text matching tasks, such as classification, visual question answering, multimodal retrieval, and visual grounding.

One of the core limitations of both CLIP-based models and MLLM embeddings is that contrastive learning tends to focus on global semantic information while being insensitive to fine-grained differences \cite{zhong2022regionclip,li2022grounded,wang2024granular,wang2025advancing,wang2025cof,yu2024texthawk,maninis2024tips,naeem2024silc,jing2024fineclip,shi2025catching,jiang2025analyzing,jiang2025exploring,yang2025new,hou2025fire,xiao2025flair,xie2025fg}. Although for MLLM embeddings, we can add prompts such as ``Represent this type of fine-grained information: \dots'' to provide explicit guidance, this is not a universal solution. As shown in \cref{fig:global-or-fine-grained}, due to the diversity of image-text-to-image-text matching tasks, whether the query and target should focus on global or fine-grained semantic information varies case by case. For a photo of a dog or a brief textual response, global semantic information is sufficient, whereas for images containing multiple objects or screenshots of posters, the recognition of fine-grained semantic information becomes particularly critical. Therefore, we claim that \textbf{adaptively integrating global and fine-grained perception is a key issue for MLLM embeddings, rather than focusing solely on fine-grained perception.}

Some CLIP-based works make minor modifications to image-caption pairs, such as ``a black cat wearing a yellow bow tie'' and ``a black cat wearing a red bow tie'' to construct hard negative samples and improve the model's fine-grained understanding \cite{yuksekgonul2022and,patel2024tripletclip,wang2025vitrix,zhao2024ultraedit}. However, manually identifying fine-grained elements and performing large-scale edits is challenging and costly, especially for long-sequence image-text mixed data in MLLM embeddings. A common solution is to identify hard negative samples from existing training batches and amplify the weight of these data points \cite{he2020momentum,chen2020simple,meng2024sfrembedding,lee2024nv,lan2025llave,thirukovalluru2025breaking}. Recently, \citet{xue2025improve} deeply explore the mechanism of negative samples in backpropagation and explicitly amplify their gradient contributions based on their difficulty. However, only smooth contrastive losses are compatible with explicit gradient amplification techniques, while sharp contrastive losses like MetaEmbed \cite{xiao2025metaembed} are not suitable.

In this paper, we propose \textbf{A}daptive \textbf{G}lobal and \textbf{F}ine-grained perceptual \textbf{F}usion for MLLM \textbf{Embed}dings (\model). In addition to the original embedding used for global perception, \model~employs learnable prompt tokens to guide the MLLM to spontaneously focus on fine-grained information across different dimensions and generate corresponding embeddings. To adapt to various cases in image-text–to-image-text matching tasks, we compute four types of similarities: global-to-global, fine-grained-to-global, global-to-fine-grained, and fine-grained-to-fine-grained. These similarities are aggregated using smoothed $\operatorname{logsumexp}$, which aims to enable the MLLM to softly favor more suitable patterns. We theoretically demonstrate that this smooth and adaptive fusion approach is compatible with the Explicit Gradient Amplification (EGA) \cite{xue2025improve} technique, thereby achieving hard negative enhancement without the need for additional datasets.

In summary, our contributions are as follows: (1) We propose \model, a novel multimodal embedding framework that can adaptively integrate global and fine-grained information while being compatible with the EGA technique. (2) We guide the MLLM to generate global and fine-grained embeddings based on learnable prompt tokens, and appropriately aggregate their similarities using $\operatorname{logsumexp}$. (3) We conduct an in-depth analysis of the gradient expansion for \model~and theoretically derive the form of its gradient amplification for hard negatives. (4) The results on the MMEB and MMVP-VLM benchmarks demonstrate the outstanding performance of \model~in both general and fine‑grained understanding, while the ablation study further validates the necessity of each module.

\section{Related Work}

\begin{figure*}[t]
  \begin{center}
    \centerline{\includegraphics[width=\textwidth]{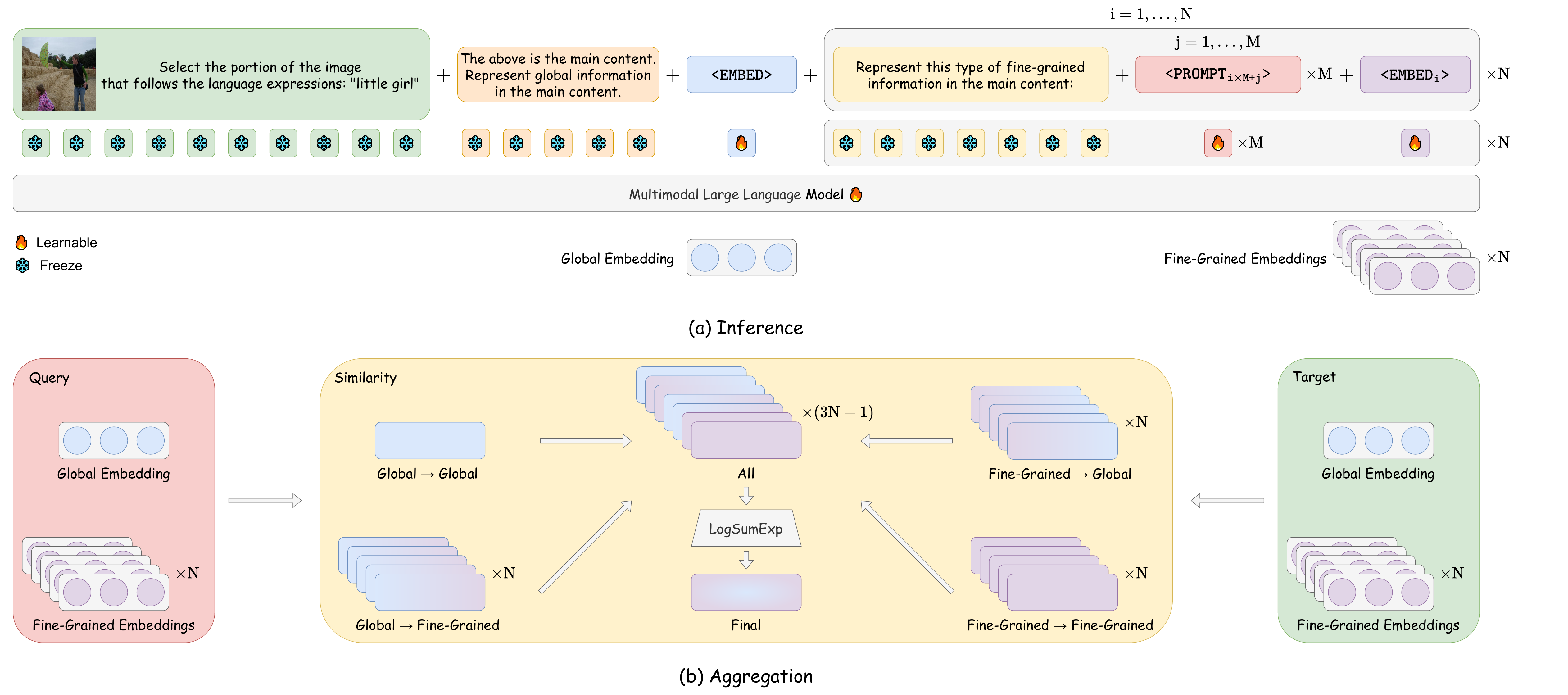}}
    \caption{
      Framework of \model. (a): MLLM generates a global embedding and $N$ fine-grained embeddings. (b): Four types of similarity corresponding to four perception patterns are computed and aggregated via $\operatorname{logsumexp}$.
    }
    \label{fig:framework}
  \end{center}
  \vskip -0.2in
\end{figure*}

\paragraph{CLIP-based embedding models.}

CLIP \cite{radford2021learning} employs a dual-tower architecture with separate image and text encoders to generate corresponding embeddings. It performs contrastive learning on a vast collection of image-caption pairs from the web, achieving zero-shot performance on downstream tasks comparable to fully supervised models. Subsequent works further mature the CLIP framework. MetaCLIP \cite{xu2024demystifying} systematically constructs CLIP’s training data and surpasses its performance. EVA-CLIP \cite{sun2023eva} and EVA-CLIP-18B \cite{sun2024eva} scale the parameters of CLIP to 8B and 18B respectively, further pushing the limits of its capabilities. SigLIP \cite{zhai2023sigmoid} replaces the $\operatorname{softmax}$ normalization in CLIP with a $\operatorname{sigmoid}$ loss, decoupling batch size from the loss function, while SigLIP-2 \cite{tschannen2025siglip} builds upon SigLIP by integrating and forming a comprehensive training pipeline.

The limitation of CLIP lies in its focus on global semantic perception while overlooking fine-grained information. NegCLIP \cite{yuksekgonul2022and}, TripletCLIP \cite{patel2024tripletclip}, and CLIP-IN \cite{wang2025vitrix} construct hard negatives with subtle differences and incorporate them into contrastive learning batches to guide the model in making distinctions. EqSim \cite{wang2023equivariant} introduces equivariant regularization for similar positive and negative sample pairs, making their relative positions in the embedding space more reasonable. DAC \cite{doveh2023dense} generates additional captions based on LLM-Expander and SAM-Expander, and employs a negative generator to create hard negatives for them. MosaiCLIP \cite{singh2023coarse} generates scene graphs for images and captions separately, and perturbs their nodes to construct hard negative sub-graphs. CECLIP \cite{zhang2024contrasting} generates hard negatives specifically for different attributes within captions. Additionally, RegionCLIP \cite{zhong2022regionclip}, GLIP \cite{li2022grounded}, CLIPSelf \cite{wu2024clipself}, SILC \cite{naeem2024silc}, FineCLIP \cite{jing2024fineclip}, and FGCLIP \cite{xie2025fg} achieve fine-grained image-text alignment from the global to the regional level.

\paragraph{MLLM-based embedding models.}

VLM2Vec \cite{jiang2025vlm2vec} and E5-V \cite{jiang2024e5} make full use of the powerful capabilities of MLLMs to extend the input of embedding models from single images or texts to mixed image-text content. They extract the final hidden state of the last token in the MLLM input as the embedding and perform contrastive learning on their proposed MMEB dataset. Furthermore, GME \cite{zhang2024gme} and VLM2Vec-V2 \cite{meng2025vlm2vec} introduce additional modalities such as visual documents and videos. MegaPairs \cite{zhou2025megapairs} and mmE5 \cite{chen2025mme5} synthesize high-quality training data for MLLM embeddings. MoCA \cite{chen2025moca} proposes a two-stage training framework for transforming MLLMs into embedding models, which involves modality-aware continual pre-training and heterogeneous contrastive fine-tuning. Recently, MetaEmbed \cite{xiao2025metaembed} introduces a multi-vector retrieval mechanism, making retrieval costs scalable.

Similar to CLIP, a key point for improving the training efficiency and quality of MLLM embeddings lies in fully leveraging hard negatives. UniME \cite{gu2025breaking} filters false negatives and samples hard negatives for each batch during the instruction tuning stage. B3 \cite{thirukovalluru2025breaking} constructs batches rich in hard negatives based on the ranking from a teacher model and METIS clustering. According to the similarity between negative samples and queries, i.e., the difficulty level, LLaVE \cite{lan2025llave} and QQMM \cite{xue2025improve} weight the corresponding terms in the contrastive loss and the corresponding gradients during backpropagation, respectively.

\section{Method}

\subsection{Adaptive Global and Fine-Grained Perceptual Fusion}

To address the different cases presented in \cref{fig:global-or-fine-grained}, we drive the MLLM to adaptively generate and aggregate global and fine‑grained embeddings, with its framework illustrated in \cref{fig:framework}. We append the prompt ``The above is the main content. Represent global information in the main content.'' after the input content to clearly delimit the scope for embedding and guide the MLLM to focus first on global information. Subsequently, similar to VLM2Vec \cite{jiang2025vlm2vec}, we use a special token $\mathtt{{<}EMBED{>}}$ to indicate the generation position of the global embedding $\mathbf{x}_0 \in \mathbb{R}^D$. For further extraction of fine-grained semantic information, we introduce $N$ parallel fine-grained embedding modules after the global embedding module. Each of these modules consists of three components: first, a prompt ``Represent this type of fine-grained information in the main content:'' to explicitly guide the MLLM; second, $M$ learnable $\mathtt{{<}PROMPT_{i \times M + j}{>}}$ tokens, where $i=1,\dots,N$ denotes the module index and $j=1,\dots,M$ denotes the prompt index, aiming to allow the MLLM to autonomously determine the types of fine-grained information to focus on; and finally, a special $\mathtt{{<}EMBED_i{>}}$ token to mark the position where the fine‑grained embedding $\mathbf{x}_i \in \mathbb{R}^D$ of this module is generated. Thus, all fine-grained modules collectively produce $N$ embeddings $\{ \mathbf{x}_i \}_{i=1}^N$, each focusing on different fine-grained semantics.

Now we have the global embedding $\mathbf{x}^q_0$ and fine-grained embeddings $\{\mathbf{x}^q_i\}_{i=1}^N$ for the query, as well as the global embedding $\mathbf{x}^t_0$ and fine-grained embeddings $\{\mathbf{x}^t_i\}_{i=1}^N$ for the target. Corresponding to the four cases in \cref{fig:global-or-fine-grained}, we compute the following four types of similarity:
\begin{equation}
    \label{eq:similarity-fg}
    \begin{cases}
        s^\mathrm{g2g} \left( X^q, X^t \right) = \mathbf{x}^q_0 \cdot \mathbf{x}^t_0,\\
        s^\mathrm{f2g}_i \left( X^q, X^t \right) = \mathbf{x}^q_i \cdot \mathbf{x}^t_0, & i = 1, \dots, N, \\
        s^\mathrm{g2f}_i \left( X^q, X^t \right) = \mathbf{x}^q_0 \cdot \mathbf{x}^t_i, & i = 1, \dots, N, \\
        s^\mathrm{f2f}_i \left( X^q, X^t \right) = \mathbf{x}^q_i \cdot \mathbf{x}^t_i, & i = 1, \dots, N,
    \end{cases}
\end{equation}
where $X^q = \left[ \mathbf{x}^q_0, \mathbf{x}^q_1, \dots, \mathbf{x}^q_N \right]^\top \in \mathbb{R}^{(N+1) \times D}$ and $X^t = \left[ \mathbf{x}^t_0, \mathbf{x}^t_1, \dots, \mathbf{x}^t_N \right]^\top \in \mathbb{R}^{(N+1) \times D}$ are the complete embeddings for the query and target, respectively.

For a given positive pair, not all types of similarity align with its perceptual pattern, but the more compatible similarities should have higher values. An intuitive aggregation method is to take the maximum of all $(3N+1)$ similarities \cite{xiao2025metaembed}. However, perceptual patterns are often not singular, requiring simultaneous attention to multiple global or fine-grained information, with varying degrees of emphasis. Moreover, the maximum function is sharp and lacks favorable gradient properties, making it unsuitable for explicit gradient amplification techniques to enhance hard negatives. Therefore, we adopt the $\operatorname{logsumexp}$ function for similarity aggregation, which, as a smooth approximation of the maximum, can adaptively capture multiple dominant perceptual patterns while preserving the differentiability and gradient stability of the loss function:
\begin{multline}
    \label{eq:similarity-final}
    s^\mathrm{final} = \log \left[ \exp \left( s^\mathrm{g2g} \right) + \sum_{i=1}^N \exp \left( s^\mathrm{f2g}_i \right) \right. \\
    \left. + \sum_{i=1}^N \exp \left( s^\mathrm{g2f}_i \right) + \sum_{i=1}^N \exp \left( s^\mathrm{f2f}_i \right) \right].
\end{multline}
Then we obtain the InfoNCE loss for training:
\begin{equation}
    \label{eq:infonce}
    \resizebox{\columnwidth}{!}{$
    \mathcal{L} =
    -\log \frac{\exp \left[ \phi \left( X^q, X^{t^+} \right) \right]}{\exp \left[ \phi \left( X^q, X^{t^+} \right) \right] + \sum_{i=1}^{B-1} \exp \left[ \phi \left( X^q, X^{t^-_i} \right) \right]},
    $}
\end{equation}
where $\phi(X^q, X^t) = \frac{1}{\tau} s^\mathrm{final}(X^q, X^t)$, $\tau$ is the temperature hyperparameter, $B$ is the batch size, $t^+$ is the positive target, and $t^-_i$ is the $i$-th negative target.

\subsection{Explicit Hard Negative Gradient Amplification for \model}
\label{sec:ega}

A larger batch size can effectively improve the quality of contrastive learning by introducing richer negative samples. However, GPU memory often limit the increase in batch size, a challenge that is particularly pronounced for MLLM embeddings, as their inputs typically include multiple images and long text sequences. To address this bottleneck, VLM2Vec \cite{jiang2025vlm2vec} employs the GradCache technique \cite{gao2021scaling} during backpropagation, decomposing the gradient of the loss function $\mathcal{L}$ with respect to the MLLM parameters $\Theta$ according to the chain rule:
\begin{equation}
    \frac{\partial \mathcal{L}}{\partial \Theta} = \frac{\partial \mathcal{L}}{\partial X^q} \frac{\partial X^q}{\partial \Theta} + \frac{\partial \mathcal{L}}{\partial X^{t^+}} \frac{\partial X^{t^+}}{\partial \Theta} + \sum_{i=1}^{B-1} \frac{\partial \mathcal{L}}{\partial X^{t_i^-}} \frac{\partial X^{t_i^-}}{\partial \Theta}.
\end{equation}
In practice, the gradients of the loss function with respect to each embedding $\left\{ \frac{\partial \mathcal{L}}{\partial X^q}, \frac{\partial \mathcal{L}}{\partial X^{t^+}}, \frac{\partial \mathcal{L}}{\partial X^{t_0^-}}, \dots, \frac{\partial \mathcal{L}}{\partial X^{t_{B-1}^-}} \right\}$ are first computed separately by PyTorch's automatic differentiation and cached after detaching. Then, these gradients are dot-multiplied with their corresponding embeddings and backpropagated, which can be parallelized by splitting the full batch into smaller sub-batches.

Before delving into the analysis of cached gradients, we first denote the normalized probabilities of $s^\mathrm{g2g}, s_i^\mathrm{f2g}, s_i^\mathrm{g2f}, s_i^\mathrm{f2f}$ as $w^\mathrm{g2g}, w_i^\mathrm{f2g}, w_i^\mathrm{g2f}, w_i^\mathrm{f2f}$, respectively. For example:
\begin{equation}
    \label{eq:norm-prob-part}
    \resizebox{\columnwidth}{!}{$
        w^\mathrm{g2g} = \frac{\exp \left( s^\mathrm{g2g} \right)}{\exp \left( s^\mathrm{g2g} \right) + \sum_{j=1}^N \exp \left( s_j^\mathrm{f2g} \right) + \sum_{j=1}^N \exp \left( s_j^\mathrm{g2f} \right) + \sum_{j=1}^N \exp \left( s_j^\mathrm{f2f} \right)}.
    $}
\end{equation}
Additionally, the classification probabilities of the target samples are denoted as:
\begin{equation}
    \label{eq:cls-prob}
    \begin{cases}
        p^+ = \frac{\exp \left[ \phi \left( X^q, X^{t^+} \right) \right]}{\exp \left[ \phi \left( X^q, X^{t^+} \right) \right] + \sum_{j=1}^{B-1} \exp \left[ \phi \left( X^q, X^{t^-_j} \right) \right]}, \\
        p_i^- = \frac{\exp \left[ \phi \left( X^q, X^{t_i^-} \right) \right]}{\exp \left[ \phi \left( X^q, X^{t^+} \right) \right] + \sum_{j=1}^{B-1} \exp \left[ \phi \left( X^q, X^{t^-_j} \right) \right]}.
    \end{cases}
\end{equation}
Then, for \model, we expand $\frac{\partial \mathcal{L}}{\partial X^q}$ based on \cref{eq:similarity-fg,eq:similarity-final,eq:infonce}:
\begin{equation}
    \label{eq:gradient-q}
    \begin{cases}
        \frac{\partial \mathcal{L}}{\partial \mathbf{x}_0^q} = & \frac{1}{\tau} \sum_{i=1}^{B-1} p_i^- \left[ w^\mathrm{g2g} \left( X^q, X^{t_i^-} \right) \mathbf{x}_0^{t_i^-} \right. \\
        & + \sum_{j=1}^N w^\mathrm{g2f}_j \left( X^q, X^{t_i^-} \right) \mathbf{x}_j^{t_i^-} \\
        & - w^\mathrm{g2g} \left( X^q, X^{t^+} \right) \mathbf{x}_0^{t^+} \\
        & \left. - \sum_{j=1}^N w_j^\mathrm{g2f} \left( X^q, X^{t^+} \right) \mathbf{x}_j^{t^+} \right], \\
        \frac{\partial \mathcal{L}}{\partial \mathbf{x}_j^q} = & \frac{1}{\tau} \sum_{i=1}^{B-1} p_i^- \left[ w_j^\mathrm{f2g} \left( X^q, X^{t_i^-} \right) \mathbf{x}_0^{t_i^-} \right. \\
        & + w_j^\mathrm{f2f} \left( X^q, X^{t_i^-} \right) \mathbf{x}_j^{t_i^-} \\
        & - w_j^\mathrm{f2g} \left( X^q, X^{t^+} \right) \mathbf{x}_0^{t^+} \\
        & \left. - w_j^\mathrm{f2f} \left( X^q, X^{t^+} \right) \mathbf{x}_j^{t^+} \right], \ j = 1, \dots, N.
    \end{cases}
\end{equation}
Similarly, $\frac{\partial \mathcal{L}}{\partial X^{t^+}}$ is derived as:
\begin{equation}
    \label{eq:gradient-t+}
    \begin{cases}
        \frac{\partial \mathcal{L}}{\partial \mathbf{x}_0^{t^+}} = & \frac{1}{\tau} \left( p^+ - 1 \right) \left[ w^\mathrm{g2g} \left( X^q, X^{t^+} \right) \mathbf{x}^q_0 \right. \\
        & \left. + \sum_{j=1}^N w^\mathrm{f2g}_j \left( X^q, X^{t^+} \right) \mathbf{x}^q_j \right], \\
        \frac{\partial \mathcal{L}}{\partial \mathbf{x}_j^{t^+}} = & \frac{1}{\tau} \left (p^+ - 1 \right) \left[ w^\mathrm{g2f}_j \left( X^q, X^{t^+} \right) \mathbf{x}_0^q \right. \\
        & \left. + w^\mathrm{f2f}_j \left( X^q, X^{t^+} \right) \mathbf{x}^q_j \right], \ j = 1, \dots, N.
    \end{cases}
\end{equation}
Finally, $\frac{\partial \mathcal{L}}{\partial X^{t_i^-}}$ can be expressed as:
\begin{equation}
    \label{eq:gradient-t-}
    \begin{cases}
        \frac{\partial \mathcal{L}}{\partial \mathbf{x}_0^{t_i^-}} = & \frac{1}{\tau} p_i^- \left[ w^\mathrm{g2g} \left( X^q, X^{t_i^-} \right) \mathbf{x}_0^q \right. \\
        & \left. + \sum_{j=1}^N w^\mathrm{f2g}_j \left( X^q, X^{t_i^-} \right) \mathbf{x}_j^q \right], \\
        \frac{\partial \mathcal{L}}{\partial \mathbf{x}_j^{t_i^-}} = & \frac{1}{\tau} p_i^- \left[ w^\mathrm{g2f}_j \left( X^q, X^{t_i^-} \right) \mathbf{x}_0^q \right. \\
        & \left. + w^\mathrm{f2f}_j \left( X^q, X^{t_i^-} \right) \mathbf{x}_j^q \right], \ j = 1, \dots, N.
    \end{cases}
\end{equation}
For detailed derivation, please refer to Appendix \ref{sec:detailed derivation}. The following conclusions can be drawn from \cref{eq:gradient-q,eq:gradient-t+,eq:gradient-t-}: (1) Negative targets with higher classification probabilities (i.e., harder negatives) contribute more to the gradient. (2) Similarities with higher normalized weights (i.e., more aligned perceptual patterns) contribute more to the gradient. It should be noted that the above metrics rely on the embeddings generated by the MLLM itself. Since the pre-trained MLLM has already distinguished a large number of positive and negative targets, its assessment of the hardness of negative targets is relatively reliable. However, its understanding of global or fine-grained perceptual patterns is learned from scratch, making it difficult to accurately identify dominant perceptual patterns in the early stages of training. Therefore, at present, we only perform Explicit Gradient Amplification (EGA) \cite{xue2025improve} for hard negatives, reserving its application to perceptual patterns for future research as \model~further matures and scales up.

Similar to \citet{xue2025improve}, we define the hardness of a negative target based on its similarity difference from the positive target:
\begin{equation}
    \label{eq:hardness}
    h_i^- = \exp \left\{ \alpha \cdot \left[ s^\mathrm{final} \left( X^q, X^{t_i^-} \right) - s^\mathrm{final} \left( X^q, X^{t^+} \right) \right] \right\},
\end{equation}
where $\alpha$ is a hyperparameter that controls the amplification magnitude. Then, the classification probabilities of negative targets are reassigned according to hardness:
\begin{equation}
    \hat{p}_i^- = p_i^- \cdot h_i^-, \quad \bar{p}_i^- = \frac{\hat{p}_i^-}{\sum_{i=1}^{B-1} \hat{p}_i^-} \cdot \sum_{i=1}^{B-1} p_i^-.
\end{equation}
Substituting the updated probabilities $\{\bar{p}_i^-\}_{i=1}^{B-1}$ back into \cref{eq:gradient-q,eq:gradient-t+,eq:gradient-t-} yields cached gradients amplified by hard negatives.

\section{Experiment}

\begin{table*}[t]
  \caption{Results on the MMEB benchmark. Results for CLIP, BLIP2, MagicLens, SigLIP, OpenCLIP, UniIR, E5-V, and VLM2Vec are from \citet{jiang2025vlm2vec}; results for EVA-CLIP, LLaVE, UniME, B3, and QQMM are from \citet{xue2025improve}; and results for UniME-V2, UME-R1, and ReMatch are from their original papers. For detailed results, please refer to \cref{tab:mmeb-detailed} in Appendix \ref{sec:mmeb-detailed}.}
  \label{tab:mmeb}
  \begin{center}
    \resizebox{\textwidth}{!}{
    \begin{tabular}{lcccccccc}
      \toprule
      \multirow{2}{*}{\textbf{Model}} & \multirow{2}{*}{\textbf{Size}} & \multicolumn{4}{c}{\textbf{Per Meta-Task Score}} & \multicolumn{3}{c}{\textbf{Average Score}} \\
      \cmidrule(lr){3-6} \cmidrule(lr){7-9}
      & & \textbf{Classification} & \textbf{VQA} & \textbf{Retrieval} & \textbf{Grounding} & \textbf{IND} & \textbf{OOD} & \textbf{Overall} \\
      \midrule
      \textbf{\# of datasets} & & 10 & 10 & 12 & 4 & 20 & 16 & 36 \\
      \midrule
      \multicolumn{9}{c}{\textit{CLIP-Based Embedding Models}} \\
      \midrule
      CLIP \cite{radford2021learning} & 0.4B & 42.8 & 9.1 & 53.0 & 51.8 & 37.1 & 38.7 & 37.8 \\
      BLIP2 \cite{li2023blip} & 1.2B & 27.0 & 4.2 & 33.9 & 47.0 & 25.3 & 25.1 & 25.2 \\
      MagicLens \cite{zhang2024magiclens} & 0.4B & 38.8 & 8.3 & 35.4 & 26.0 & 31.0 & 23.7 & 27.8 \\
      SigLIP \cite{zhai2023sigmoid} & 0.9B & 40.3 & 8.4 & 31.6 & 59.5 & 32.3 & 38.0 & 34.8 \\
      OpenCLIP \cite{cherti2023reproducible} & 0.4B & 47.8 & 10.9 & 52.3 & 53.3 & 39.3 & 40.2 & 39.7 \\
      UniIR ($\text{BLIP}_\text{FF}$) \cite{wei2024uniir} & unknown & 42.1 & 15.0 & 60.1 & 62.2 & 44.7 & 40.4 & 42.8 \\
      EVA-CLIP \cite{sun2024eva} & 8B & 56.0 & 10.4 & 49.2 & 58.9 & 38.1 & 45.6 & 43.7 \\
      UniIR ($\text{CLIP}_\text{SF}$) \cite{wei2024uniir} & unknown & 44.3 & 16.2 & 61.8 & 65.3 & 47.1 & 41.7 & 44.7 \\
      \midrule
      \multicolumn{9}{c}{\textit{MLLM-Based Embedding Models}} \\
      \midrule
      E5-V \cite{jiang2024e5} & 7B & 21.8 & 4.9 & 11.5 & 19.0 & 14.9 & 11.5 & 13.3 \\
      VLM2Vec \cite{jiang2025vlm2vec} & 7B & 61.2 & 49.9 & 67.4 & 86.1 & 67.5 & 57.1 & 62.9 \\
      LLaVE \cite{lan2025llave} & 7B & 65.7 & 65.4 & 70.9 & 91.9 & 75.0 & 64.4 & 70.3 \\
      UniME \cite{gu2025breaking} & 7B & 66.8 & 66.6 & 70.5 & 90.9 & 74.6 & 65.8 & 70.7 \\
      UniME-V2 \cite{gu2025unime} & 7B & 65.3 & 67.6 & 72.9 & 90.2 & 74.8 & 66.7 & 71.2 \\
      UME-R1 \cite{lan2025ume} & 7B & 67.1 & 69.2 & 71.9 & 84.9 & 76.1 & 65.1 & 71.3 \\
      B3 \cite{thirukovalluru2025breaking} & 7B & 70.0 & 66.5 & \textbf{74.1} & 84.6 & 75.9 & 67.1 & 72.0 \\
      QQMM \cite{xue2025improve} & 7B & 69.9 & 70.0 & 72.1 & 86.0 & 77.2 & 66.6 & 72.5 \\
      ReMatch \cite{liu2025rematch} & 7B & 65.8 & \textbf{73.6} & \textbf{74.1} & \textbf{92.5} & 78.1 & 68.2 & 73.7 \\
      \model~(Ours) & 7B & \textbf{72.6} & 72.6 & 73.8 & 90.0 & \textbf{79.5} & \textbf{69.1} & \textbf{74.9} \\
      \bottomrule
    \end{tabular}
    }
  \end{center}
  \vskip -0.1in
\end{table*}

\begin{table*}[t]
  \caption{Zero-shot performance on the MMVP-VLM benchmark. The attributes of the $9$ subsets are abbreviated as follows: Orientation (Ori.), Presence (Pre.), State (Sta.), Quantity (Qua.), Spatial (Spa.), Color (Col.), Structural Character (Str.), Text (Tex.), and Camera Perspective (Cam.). Results for CLIP, DIVA, DFN, SigLIP2, and CLIP-IN are from \citet{wang2025vitrix}; and results for LLaVE, UniME, UniME-V2, UME-R1, and QQMM are from our reproduction.}
  \label{tab:mmvp}
  \begin{center}
    \resizebox{\textwidth}{!}{
    \begin{tabular}{lccccccccccc}
      \toprule
      \textbf{Model} & \textbf{Size} & \textbf{Ori.} & \textbf{Pre.} & \textbf{Sta.} & \textbf{Qua.} & \textbf{Spa.} & \textbf{Col.} & \textbf{Str.} & \textbf{Tex.} & \textbf{Cam.} & \textbf{Average} \\
      \midrule
      \multicolumn{12}{c}{\textit{CLIP-Based Embedding Models}} \\
      \midrule
      CLIP \cite{radford2021learning} & 0.4B & 0.0 & 20.0 & 40.0 & 20.0 & 6.7 & 20.0 & 33.3 & 6.7 & 40.0 & 20.0 \\
      DIVA \cite{wang2024diffusion} & 0.4B & 26.7 & 20.0 & 33.3 & 13.3 & 13.3 & 46.7 & 26.7 & 6.7 & 40.0 & 25.2 \\
      DFN \cite{fang2023data} & 1B & 20.0 & 26.7 & \textbf{73.3} & 26.7 & 26.7 & 66.7 & 46.7 & 20.0 & 53.3 & 39.9 \\
      SigLIP2 \cite{tschannen2025siglip} & 0.9B & 13.3 & 20.0 & 60.0 & 26.7 & 6.7 & 80.0 & 53.3 & 20.0 & 40.0 & 35.6 \\
      CLIP-IN \cite{wang2025vitrix} & 1B & 20.0 & 26.7 & \textbf{73.3} & 26.7 & 33.3 & 66.7 & 46.7 & 26.7 & 53.3 & 41.5 \\
      \midrule
      \multicolumn{12}{c}{\textit{MLLM-Based Embedding Models}} \\
      \midrule
      LLaVE \cite{lan2025llave} & 7B & 53.3 & 40.0 & 66.7 & 63.3 & 43.3 & 73.3 & 53.3 & 30.0 & \textbf{60.0} & 53.7 \\
      UniME \cite{gu2025breaking} & 7B & \textbf{60.0} & 43.3 & \textbf{73.3} & 63.3 & \textbf{56.7} & 76.7 & 60.0 & 43.3 & 50.0 & 58.5 \\
      UniME-V2 \cite{gu2025unime} & 7B & 46.7 & \textbf{46.7} & \textbf{73.3} & 56.7 & 50.0 & 73.3 & 63.3 & 43.3 & \textbf{60.0} & 57.0 \\
      UME-R1 \cite{lan2025ume} & 7B & 53.3 & \textbf{46.7} & 66.7 & 60.0 & 43.3 & \textbf{83.3} & 53.3 & 40.0 & \textbf{60.0} & 56.3 \\
      QQMM \cite{xue2025improve} & 7B & 53.3 & 43.3 & \textbf{73.3} & 46.7 & 36.7 & \textbf{83.3} & 50.0 & 43.3 & 56.7 & 54.1 \\
      \model~(Ours) & 7B & \textbf{60.0} & 43.3 & \textbf{73.3} & \textbf{73.3} & 46.7 & \textbf{83.3} & \textbf{66.7} & \textbf{50.0} & 56.7 & \textbf{61.5} \\
      \bottomrule
    \end{tabular}
    }
  \end{center}
  \vskip -0.1in
\end{table*}

\subsection{Training and Evaluation on MMEB}
\label{sec:mmeb}

The MMEB (Massive Multimodal Embedding Benchmark) \cite{jiang2025vlm2vec} is a comprehensive benchmark designed to holistically evaluate the capabilities of vision-language models across diverse tasks and data distributions. It comprises $36$ datasets aggregated into four core meta-tasks: classification ($10$ datasets), visual question answering (VQA, $10$ datasets), retrieval ($12$ datasets), and visual grounding ($4$ datasets). A key feature of MMEB is its explicit evaluation of model generalization by partitioning tasks into in-distribution (IND, $20$ datasets) for training and out-of-distribution (OOD, $16$ datasets) for evaluation. This structure provides a rigorous testbed for assessing a model's core competency across different visual-language understanding paradigms and its robustness to distributional shifts.

We fine-tune our \model~using the MMEB-train dataset based on the pretrained QQMM \cite{xue2025improve}, which is also trained solely on the MMEB-train dataset without external data. \model~incorporates $N=10$ fine-grained embedding modules, each equipped with $M=10$ prompt tokens. Training is conducted for $40$ steps with a batch size of $1024$. In the MLLM, both the vision encoder and the projection layer are frozen, while the LLM is fine-tuned using LoRA with a rank of $8$, a scaling factor of $16$, a dropout rate of $0.05$, and an initial learning rate of $5 \times 10^{-5}$. The learnable special tokens, including prompt tokens and embedding tokens, are assigned an initial learning rate of $0.05$. The temperature $\tau$ and the amplification magnitude $\alpha$ are set to $0.02$ and $20$, respectively. Additional experimental results regarding the impact of hyperparameters are provided in Appendix \ref{sec:hyper}.

For fairness, we compare our \model~with multimodal embedding models that are also trained solely on the MMEB-train dataset. As shown in Table 1 (see \cref{tab:mmeb-detailed} in Appendix \ref{sec:mmeb-detailed} for detailed results), AGFF-Embed achieves an overall score of $74.9\%$, surpassing all CLIP-based embedding models and MLLM-based embedding models with comparable model sizes. Specifically, compared to QQMM \cite{xue2025improve} before fine-tuning, \model~improves the overall score by $2.4\%$. Furthermore, on both the IND and OOD subsets, \model~achieves state-of-the-art performance, demonstrating strong generalization capabilities. Analysis of the time costs for training and inference can be found in Appendix \ref{sec:time}.

\subsection{Zero-Shot Fine-Grained Perception Evaluation}
\label{sec:mmvp}

Next, we evaluate the fine-grained comprehension capability of our \model~based on its zero-shot performance on the MMVP-VLM (Multimodal Visual Patterns for Vision Language Models) \cite{tong2024eyes} benchmark. The MMVP-VLM benchmark consists of $9$ subsets, each containing hard negatives with subtle edits targeting specific attributes, which can be easily distinguished by humans but are challenging for VLMs. These specific attributes include: Orientation (Ori.), Presence (Pre.), State (Sta.), Quantity (Qua.), Spatial (Spa.), Color (Col.), Structural Character (Str.), Text (Tex.), and Camera Perspective (Cam.). Unlike generic multimodal retrieval tasks, the MMVP-VLM benchmark is designed to systematically evaluate the potential of VLMs to go beyond surface-level object recognition and achieve complex perceptual patterns and deep semantic understanding.

We endeavor to reproduce the zero-shot performance of recently open-sourced advanced MLLM-based embedding models on the MMVP-VLM benchmark, and compared them with our \model~in \cref{tab:mmvp}. It is reasonable that all MLLM-based embedding models consistently outperform CLIP-based embedding models due to the superior multimodal understanding potential of MLLMs. Building on this, \model~further achieves an average score of $61.5\%$, surpassing all baselines, including QQMM \cite{xue2025improve} before fine-tuning. Notably, on $6$ subtasks (Ori., Sta., Qua., Col., Str., Tex.), \model~achieves state-of-the-art performance in each respective category. These results demonstrate that the incorporation of fine-grained perception mechanisms unlocks the deep discriminative ability of MLLM embeddings for specific attributes.

\subsection{Ablation Study}

Finally, we investigate the necessity of each module in \model. Specifically, we will verify the following points: (1) Is the added perceptual pattern necessary? (2) Is the compatible EGA technique necessary? (3) Is the smooth $\operatorname{logsumexp}$ similarity aggregation necessary? We will conduct ablation studies on the MMEB and MMVP-VLM benchmarks to answer these questions.

\paragraph{Is the added perceptual pattern necessary?}

As shown in \cref{eq:similarity-fg,eq:similarity-final}, compared to traditional MLLM embeddings, our \model~additionally incorporates fine-grained-to-global similarities $\{ s_i^\mathrm{f2g} \}_{i=1}^N$, global-to-fine‑grained similarities $\{ s_i^\mathrm{g2f} \}_{i=1}^N$, and fine-grained-to-fine-grained similarities $\{ s_i^\mathrm{f2f} \}_{i=1}^N$. To validate their necessity, we sequentially discard one type of perceptual pattern. In practice, we set the specified category of similarities to $-\infty$, so that they vanish in the final similarity after $\operatorname{logsumexp}$ aggregation, and the corresponding normalized probabilities $w_i^\mathrm{f2g}, w_i^\mathrm{g2f}, w_i^\mathrm{f2f}$ in the amplified gradients become $0$. Other training configurations remain consistent with the full \model~for fairness.
 
As shown in \cref{tab:ablation-mmeb}, due to the richness of test scenarios in the MMEB benchmark, the exclusion of any perceptual pattern leads to a decrease in overall performance. However, when we turn to the zero-shot performance on the MMVP-VLM benchmark, as shown in \cref{tab:ablation-mmvp}, an interesting phenomenon emerges: although the average score of the full \model~surpasses that of \model~without fine-grained-to global or fine-grained-to-fine-grained perception, it is on par with \model~without global-to-fine-grained perception. This is reasonable when we delve into the characteristics of the MMVP-VLM benchmark: all queries are images containing key fine-grained information, while all targets are brief textual captions. This unique evaluation scenario places greater emphasis on the MLLM's understanding of fine-grained details in queries and global information in targets, thereby making the exclusion of global-to-fine-grained perception inconsequential to overall performance.

\paragraph{Is the compatible EGA technique necessary?}

\begin{table*}[t]
  \caption{Ablation study results on the MMEB benchmark.}
  \label{tab:ablation-mmeb}
  \begin{center}
    \resizebox{\textwidth}{!}{
    \begin{tabular}{lccccccc}
      \toprule
      \multirow{2}{*}{\textbf{Model}} & \multicolumn{4}{c}{\textbf{Per Meta-Task Score}} & \multicolumn{3}{c}{\textbf{Average Score}} \\
      \cmidrule(lr){2-5} \cmidrule(lr){6-8}
      & \textbf{Classification} & \textbf{VQA} & \textbf{Retrieval} & \textbf{Grounding} & \textbf{IND} & \textbf{OOD} & \textbf{Overall} \\
      \midrule
      \textbf{\# of datasets} & 10 & 10 & 12 & 4 & 20 & 16 & 36 \\
      \midrule
      \model~(full model) & 72.6 & \textbf{72.6} & \textbf{73.8} & 90.0 & \textbf{79.5} & \textbf{69.1} & \textbf{74.9} \\
      \midrule
      \multicolumn{8}{c}{\textit{Perceptual Pattern Exclusion}} \\
      \midrule
      w/o fine-grained $\to$ global & 72.0 & 71.9 & 71.7 & \textbf{90.1} & 78.2 & 68.5 & 73.9 \\
      w/o global $\to$ fine-grained & \textbf{72.7} & 71.7 & 72.6 & 89.8 & 78.5 & \textbf{69.1} & 74.3 \\
      w/o fine-grained $\to$ fine-grained & 71.9 & 72.0 & 72.9 & 89.2 & 78.7 & 68.6 & 74.2 \\
      \midrule
      \multicolumn{8}{c}{\textit{EGA Exclusion}} \\
      \midrule
      w/o EGA & 72.5 & 72.2 & 73.4 & 89.3 & 79.0 & \textbf{69.1} & 74.6 \\
      \midrule
      \multicolumn{8}{c}{\textit{$\operatorname{Logsumexp}$ Similarity Aggregation Exclusion}} \\
      \midrule
      $\operatorname{max}$ aggregation & 69.2 & 68.2 & 69.4 & 88.1 & 74.2 & 67.2 & 71.1 \\
      $\operatorname{mean-max}$ aggregation \cite{xiao2025metaembed} & 71.0 & 70.3 & 71.5 & 87.7 & 76.9 & 67.6 & 72.8 \\
      \bottomrule
    \end{tabular}
    }
  \end{center}
\end{table*}

\begin{table*}[t]
  \caption{Ablation study results on the MMVP-VLM benchmark.}
  \label{tab:ablation-mmvp}
  \begin{center}
    \resizebox{\textwidth}{!}{
    \begin{tabular}{lcccccccccc}
      \toprule
      \textbf{Model} & \textbf{Ori.} & \textbf{Pre.} & \textbf{Sta.} & \textbf{Qua.} & \textbf{Spa.} & \textbf{Col.} & \textbf{Str.} & \textbf{Tex.} & \textbf{Cam.} & \textbf{Average} \\
      \midrule
      \model~(full model) & 60.0 & 43.3 & 73.3 & 73.3 & 46.7 & 83.3 & \textbf{66.7} & 50.0 & 56.7 & \textbf{61.5} \\
      \midrule
      \multicolumn{11}{c}{\textit{Perceptual Pattern Exclusion}} \\
      \midrule
      w/o fine-grained $\to$ global & 46.7 & 33.3 & 73.3 & \textbf{76.7} & 46.7 & \textbf{86.7} & 50.0 & 53.3 & \textbf{63.3} & 58.9 \\
      w/o global $\to$ fine-grained & 60.0 & 46.7 & \textbf{76.7} & 73.3 & 50.0 & \textbf{86.7} & 60.0 & 40.0 & 56.7 & 61.1 \\
      w/o fine-grained $\to$ fine-grained & 60.0 & 43.3 & 70.0 & 66.7 & \textbf{53.3} & 80.0 & 60.0 & 43.3 & 60.0 & 59.6 \\
      \midrule
      \multicolumn{11}{c}{\textit{EGA Exclusion}} \\
      \midrule
      w/o EGA & \textbf{63.3} & \textbf{50.0} & 70.0 & 63.3 & \textbf{53.3} & 80.0 & 63.3 & 36.7 & 60.0 & 60.0 \\
      \midrule
      \multicolumn{11}{c}{\textit{$\operatorname{Logsumexp}$ Similarity Aggregation Exclusion}} \\
      \midrule
      $\operatorname{max}$ aggregation & 36.7 & 46.7 & 73.3 & 60.0 & 40.0 & \textbf{86.7} & 60.0 & \textbf{56.7} & 50.0 & 56.7 \\
      $\operatorname{mean-max}$ aggregation \cite{xiao2025metaembed} & 53.3 & 43.3 & 73.3 & 56.7 & 40.0 & 80.0 & 53.3 & 46.7 & 46.7 & 54.8 \\
      \bottomrule
    \end{tabular}
    }
  \end{center}
  \vskip -0.1in
\end{table*}

Explicit Gradient Amplification (EGA) \cite{xue2025improve} delves into the gradient expansion form of contrastive learning and leverages the unique GradCache mechanism of MLLM embeddings to achieve hard negative enhancement during backpropagation. We have theoretically adapted the EGA technique to our \model~in \cref{sec:ega}, and now we will evaluate its benefits through an ablation study. Specifically, we set the amplification magnitude $\alpha$ in \cref{eq:hardness} to $0$, which means the amplified gradients degenerate to the original gradients. All training configurations remain consistent with the full AGFF-Embed except for the exclusion of EGA. As shown in \cref{tab:ablation-mmeb,tab:ablation-mmvp}, on the MMEB and MMVP-VLM benchmarks, although \model~without EGA achieves relatively good performance, it still marginally lags behind the full model.

\paragraph{Is the smooth $\operatorname{logsumexp}$ similarity aggregation necessary?}

As a smooth approximation of the maximum, $\operatorname{logsumexp}$ aims to adaptively capture multiple dominant perceptual patterns. To verify the superiority of this similarity aggregation approach, we replace it with the $\operatorname{max}$ operation for comparison:
\begin{equation}
    s^\mathrm{final}_{\operatorname{max}} = \max \{ s^\mathrm{g2g}, \max_{1 \leq i \leq N} s^\mathrm{f2g}_i, \max_{1 \leq i \leq N} s^\mathrm{g2f}_i, \max_{1 \leq i \leq N} s^\mathrm{f2f}_i \},
\end{equation}
where $s^\mathrm{g2g}, \{ s^\mathrm{f2g}_i \}_{i=1}^N, \{ s^\mathrm{g2f}_i \}_{i=1}^N, \{ s^\mathrm{f2f}_i \}_{i=1}^N$ are defined by \cref{eq:similarity-fg}. Recently, MetaEmbed \cite{xiao2025metaembed} has proposed another method for multi-embedding aggregation, which we refer to as $\operatorname{mean-max}$ aggregation based on its formulation:
\begin{equation}
    s^\mathrm{final}_{\operatorname{mean-max}} \left( X^q, X^t \right) = \sum_{i=0}^N \max_{0 \leq j \leq N} \mathbf{x}^q_i \cdot \mathbf{x}^t_j,
\end{equation}
where the definitions of $X^q$ and $X^t$ are consistent with those in \cref{eq:similarity-fg}. Note that we do not directly compare our results with those reported by MetaEmbed \cite{xiao2025metaembed}, as it utilizes additional training data beyond the MMEB-train dataset. Instead, we only conduct ablation analysis focusing on its aggregation method.

Both $\operatorname{max}$ and $\operatorname{mean-max}$ aggregations lack favorable gradient properties and are not compatible with the EGA technique. Therefore, PyTorch's standard backpropagation is employed to update parameters, with other training configurations kept consistent with the full \model. As shown in \cref{tab:ablation-mmeb,tab:ablation-mmvp}, even compared to \model~without EGA, $\operatorname{max}$ and $\operatorname{mean-max}$ aggregation methods still perform significantly worse. Furthermore, if the EGA mechanism is introduced, this gap is further widened, which demonstrates the comprehensive superiority of $\operatorname{logsumexp}$ aggregation both in terms of the rationality of perceptual fusion and its compatibility with EGA.

\section{Conclusion}

In this paper, we propose the \model~framework, which aims to unlock the ability of MLLM embeddings to integrate multi-dimensional perception in complex scenarios. Leveraging the powerful language understanding of MLLMs, we guide the generation of global and fine-grained embeddings through clear prompt templates and learnable prompt tokens. Corresponding to four perception patterns, we compute their respective similarities and aggregate them via $\operatorname{logsumexp}$. This aggregation approach exhibits two distinct characteristics: it simultaneously captures multiple dominant perceptual patterns while maintaining an elegant gradient form. Exploiting the latter advantage, we adapt \model~with EGA technique to achieve hard negative enhancement within batches, eliminating the need for additional datasets. Evaluation results on the MMEB and MMVP-VLM benchmarks comprehensively demonstrate the state-of-the-art performance of \model~across both global and fine-grained understanding capabilities. The effectiveness of each module is further validated through ablation studies. In the future, we anticipate that \model~will serve as a new paradigm for MLLM embeddings.

\section*{Impact Statement}

This paper presents work whose goal is to advance the field of Machine
Learning. There are many potential societal consequences of our work, none
which we feel must be specifically highlighted here.

\nocite{langley00}

\bibliography{example_paper}
\bibliographystyle{icml2026}

\newpage
\appendix
\onecolumn
\section{Detailed Derivation of Cached Gradients}
\label{sec:detailed derivation}

To derive the final cached gradients, we first compute the derivatives of the intermediate variables. According to \cref{eq:similarity-fg}, we have:
\begin{equation}
    \frac{\partial s^\mathrm{g2g} \left( X^q, X^t \right)}{\partial \mathbf{x}_0^q} = \mathbf{x}_0^t, \quad \frac{\partial s^\mathrm{g2g} \left( X^q, X^t \right)}{\partial \mathbf{x}_0^t} = \mathbf{x}_0^q,
\end{equation}
\begin{equation}
    \frac{\partial s^\mathrm{f2g}_i \left( X^q, X^t \right)}{\partial \mathbf{x}_i^q} = \mathbf{x}_0^t, \quad \frac{\partial s^\mathrm{f2g}_i \left( X^q, X^t \right)}{\partial \mathbf{x}_0^t} = \mathbf{x}_i^q, \quad i = 1, \dots, N,
\end{equation}
\begin{equation}
    \frac{\partial s^\mathrm{g2f}_i \left( X^q, X^t \right)}{\partial \mathbf{x}_0^q} = \mathbf{x}_i^t, \quad \frac{\partial s^\mathrm{g2f}_i \left( X^q, X^t \right)}{\partial \mathbf{x}_i^t} = \mathbf{x}_0^q, \quad i = 1, \dots, N,
\end{equation}
\begin{equation}
    \frac{\partial s^\mathrm{f2f}_i \left( X^q, X^t \right)}{\partial \mathbf{x}_i^q} = \mathbf{x}_i^t, \quad \frac{\partial s^\mathrm{f2f}_i \left( X^q, X^t \right)}{\partial \mathbf{x}_i^t} = \mathbf{x}_i^q, \quad i = 1, \dots, N.
\end{equation}
From \cref{eq:similarity-final}, we obtain:
\begin{equation}
    \frac{\partial s^\mathrm{final}}{\partial s^\mathrm{g2g}} = \frac{\exp \left( s^\mathrm{g2g} \right)}{\exp \left( s^\mathrm{g2g} \right) + \sum_{j=1}^N \exp \left( s_j^\mathrm{f2g} \right) + \sum_{j=1}^N \exp \left( s_j^\mathrm{g2f} \right) + \sum_{j=1}^N \exp \left( s_j^\mathrm{f2f} \right)} = w^\mathrm{g2g},
\end{equation}
\begin{equation}
    \frac{\partial s^\mathrm{final}}{\partial s^\mathrm{f2g}_i} = \frac{\exp \left( s^\mathrm{f2g}_i \right)}{\exp \left( s^\mathrm{g2g} \right) + \sum_{j=1}^N \exp \left( s_j^\mathrm{f2g} \right) + \sum_{j=1}^N \exp \left( s_j^\mathrm{g2f} \right) + \sum_{j=1}^N \exp \left( s_j^\mathrm{f2f} \right)} = w^\mathrm{f2g}_i,
\end{equation}
\begin{equation}
    \frac{\partial s^\mathrm{final}}{\partial s^\mathrm{g2f}_i} = \frac{\exp \left( s^\mathrm{g2f}_i \right)}{\exp \left( s^\mathrm{g2g} \right) + \sum_{j=1}^N \exp \left( s_j^\mathrm{f2g} \right) + \sum_{j=1}^N \exp \left( s_j^\mathrm{g2f} \right) + \sum_{j=1}^N \exp \left( s_j^\mathrm{f2f} \right)} = w^\mathrm{g2f}_i,
\end{equation}
\begin{equation}
    \frac{\partial s^\mathrm{final}}{\partial s^\mathrm{f2f}_i} = \frac{\exp \left( s^\mathrm{f2f}_i \right)}{\exp \left( s^\mathrm{g2g} \right) + \sum_{j=1}^N \exp \left( s_j^\mathrm{f2g} \right) + \sum_{j=1}^N \exp \left( s_j^\mathrm{g2f} \right) + \sum_{j=1}^N \exp \left( s_j^\mathrm{f2f} \right)} = w^\mathrm{f2f}_i,
\end{equation}
which is precisely the definition of normalized probabilities as exemplified by \cref{eq:norm-prob-part}.

Noting that $\phi \left( X^q, X^t \right) = \frac{1}{\tau} s^\mathrm{final} \left( X^q, X^t \right)$, then we calculate its gradient:
\begin{equation}
    \begin{aligned}
        \frac{\partial \phi \left( X^q, X^t \right)}{\partial \mathbf{x}_0^q} &= \frac{1}{\tau} \left[ \frac{\partial s^\mathrm{final} \left( X^q, X^t \right)}{\partial s^\mathrm{g2g} \left( X^q, X^t \right)} \frac{\partial s^\mathrm{g2g} \left( X^q, X^t \right)}{\partial \mathbf{x}_0^q} + \sum_{i=1}^N \frac{\partial s^\mathrm{final} \left( X^q, X^t \right)}{\partial s^\mathrm{g2f}_i \left( X^q, X^t \right)} \frac{\partial s^\mathrm{g2f}_i \left( X^q, X^t \right)}{\partial \mathbf{x}_0^q} \right] \\
        &= \frac{1}{\tau} \left[ w^\mathrm{g2g} \left( X^q, X^t \right) \mathbf{x}_0^t + \sum_{i=1}^N w_i^\mathrm{g2f} \left(X^q, X^t \right) \mathbf{x}_i^t \right],
    \end{aligned}
\end{equation}
\begin{equation}
    \begin{aligned}
        \frac{\partial \phi \left( X^q, X^t \right)}{\partial \mathbf{x}_i^q} &= \frac{1}{\tau} \left[ \frac{\partial s^\mathrm{final} \left( X^q, X^t \right)}{\partial s^\mathrm{f2g}_i \left( X^q, X^t \right)} \frac{\partial s^\mathrm{f2g}_i \left( X^q, X^t \right)}{\partial \mathbf{x}_i^q} + \frac{\partial s^\mathrm{final} \left( X^q, X^t \right)}{\partial s^\mathrm{f2f}_i \left( X^q, X^t \right)} \frac{\partial s^\mathrm{f2f}_i \left( X^q, X^t \right)}{\partial \mathbf{x}_i^q} \right] \\
        &= \frac{1}{\tau} \left[ w^\mathrm{f2g}_i \left( X^q, X^t \right) \mathbf{x}_0^t + w_i^\mathrm{f2f} \left(X^q, X^t \right) \mathbf{x}_i^t \right], \quad i = 1, \dots, N,
    \end{aligned}
\end{equation}
\begin{equation}
    \begin{aligned}
        \frac{\partial \phi \left( X^q, X^t \right)}{\partial \mathbf{x}_0^t} &= \frac{1}{\tau} \left[ \frac{\partial s^\mathrm{final} \left( X^q, X^t \right)}{\partial s^\mathrm{g2g} \left( X^q, X^t \right)} \frac{\partial s^\mathrm{g2g} \left( X^q, X^t \right)}{\partial \mathbf{x}_0^t} + \sum_{i=1}^N \frac{\partial s^\mathrm{final} \left( X^q, X^t \right)}{\partial s^\mathrm{f2g}_i \left( X^q, X^t \right)} \frac{\partial s^\mathrm{f2g}_i \left( X^q, X^t \right)}{\partial \mathbf{x}_0^t} \right] \\
        &= \frac{1}{\tau} \left[ w^\mathrm{g2g} \left( X^q, X^t \right) \mathbf{x}_0^q + \sum_{i=1}^N w_i^\mathrm{f2g} \left(X^q, X^t \right) \mathbf{x}_i^q \right],
    \end{aligned}
\end{equation}
\begin{equation}
    \begin{aligned}
        \frac{\partial \phi \left( X^q, X^t \right)}{\partial \mathbf{x}_i^t} &= \frac{1}{\tau} \left[ \frac{\partial s^\mathrm{final} \left( X^q, X^t \right)}{\partial s^\mathrm{g2f}_i \left( X^q, X^t \right)} \frac{\partial s^\mathrm{g2f}_i \left( X^q, X^t \right)}{\partial \mathbf{x}_i^t} + \frac{\partial s^\mathrm{final} \left( X^q, X^t \right)}{\partial s^\mathrm{f2f}_i \left( X^q, X^t \right)} \frac{\partial s^\mathrm{f2f}_i \left( X^q, X^t \right)}{\partial \mathbf{x}_i^t} \right] \\
        &= \frac{1}{\tau} \left[ w^\mathrm{g2f}_i \left( X^q, X^t \right) \mathbf{x}_0^q + w_i^\mathrm{f2f} \left(X^q, X^t \right) \mathbf{x}_i^q \right], \quad i = 1, \dots, N.
    \end{aligned}
\end{equation}
We rewrite the InfoNCE loss in \cref{eq:infonce} as:
\begin{equation}
    \mathcal{L} = -\phi \left( X^q, X^{t^+} \right) + \log \left\{ \exp \left[ \phi \left( X^q, X^{t^+} \right) \right] + \sum_{i=1}^{B-1} \exp \left[ \phi \left( X^q, X^{t^-_i} \right) \right] \right\}.
\end{equation}
Finally, the cached gradients are expanded as follows:
\begin{flalign}
\resizebox{\textwidth}{!}{$
\begin{aligned}
    \frac{\partial \mathcal{L}}{\partial \mathbf{x}_0^q} &= -\frac{\partial \phi \left( X^q, X^{t^+} \right)}{\partial \mathbf{x}_0^q} + p^+ \frac{\partial \phi \left( X^q, X^{t^+} \right)}{\partial \mathbf{x}_0^q} + \sum_{i=1}^{B-1} p_i^- \frac{\partial \phi \left( X^q, X^{t_i^-} \right)}{\partial \mathbf{x}_0^q} \\
    &= \sum_{i=1}^{B-1} p_i^- \left[ \frac{\partial \phi \left( X^q, X^{t_i^-} \right)}{\partial \mathbf{x}_0^q} - \frac{\partial \phi \left( X^q, X^{t^+} \right)}{\partial \mathbf{x}_0^q} \right] \\
    &= \frac{1}{\tau} \sum_{i=1}^{B-1} p_i^- \left[ w^\mathrm{g2g} \left( X^q, X^{t_i^-} \right) \mathbf{x}_0^{t_i^-}
    + \sum_{j=1}^N w^\mathrm{g2f}_j \left( X^q, X^{t_i^-} \right) \mathbf{x}_j^{t_i^-}
    - w^\mathrm{g2g} \left( X^q, X^{t^+} \right) \mathbf{x}_0^{t^+}
    - \sum_{j=1}^N w_j^\mathrm{g2f} \left( X^q, X^{t^+} \right) \mathbf{x}_j^{t^+} \right],
\end{aligned}
$}
&&
\end{flalign}
\begin{flalign}
\resizebox{\textwidth}{!}{$
\begin{aligned}
    \frac{\partial \mathcal{L}}{\partial \mathbf{x}_j^q} &= -\frac{\partial \phi \left( X^q, X^{t^+} \right)}{\partial \mathbf{x}_j^q} + p^+ \frac{\partial \phi \left( X^q, X^{t^+} \right)}{\partial \mathbf{x}_j^q} + \sum_{i=1}^{B-1} p_i^- \frac{\partial \phi \left( X^q, X^{t_i^-} \right)}{\partial \mathbf{x}_j^q} \\
    &= \sum_{i=1}^{B-1} p_i^- \left[ \frac{\partial \phi \left( X^q, X^{t_i^-} \right)}{\partial \mathbf{x}_j^q} - \frac{\partial \phi \left( X^q, X^{t^+} \right)}{\partial \mathbf{x}_j^q} \right] \\
    &= \frac{1}{\tau} \sum_{i=1}^{B-1} p_i^- \left[ w^\mathrm{f2g}_j \left( X^q, X^{t_i^-} \right) \mathbf{x}_0^{t_i^-}
    + w^\mathrm{f2f}_j \left( X^q, X^{t_i^-} \right) \mathbf{x}_j^{t_i^-}
    - w^\mathrm{f2g}_j \left( X^q, X^{t^+} \right) \mathbf{x}_0^{t^+}
    - w^\mathrm{f2f}_j \left( X^q, X^{t^+} \right) \mathbf{x}_j^{t^+} \right], \ j=1, \dots, N,
\end{aligned}
$}
&&
\end{flalign}
\begin{flalign}
\begin{aligned}
    \frac{\partial \mathcal{L}}{\partial \mathbf{x}_0^{t^+}} &= -\frac{\partial \phi \left( X^q, X^{t^+} \right)}{\partial \mathbf{x}_0^{t^+}} + p^+ \frac{\partial \phi \left( X^q, X^{t^+} \right)}{\partial \mathbf{x}_0^{t^+}} \\
    &= \frac{1}{\tau} \left( p^+ - 1 \right) \left[ w^\mathrm{g2g} \left( X^q, X^{t^+} \right) \mathbf{x}_0^q
    + \sum_{j=1}^N w^\mathrm{f2g}_j \left( X^q, X^{t^+} \right) \mathbf{x}_j^q \right],
\end{aligned}
&&
\end{flalign}
\begin{flalign}
\begin{aligned}
    \frac{\partial \mathcal{L}}{\partial \mathbf{x}_j^{t^+}} &= -\frac{\partial \phi \left( X^q, X^{t^+} \right)}{\partial \mathbf{x}_j^{t^+}} + p^+ \frac{\partial \phi \left( X^q, X^{t^+} \right)}{\partial \mathbf{x}_j^{t^+}} \\
    &= \frac{1}{\tau} \left( p^+ - 1 \right) \left[ w^\mathrm{g2f}_j \left( X^q, X^{t^+} \right) \mathbf{x}_0^q
    + w^\mathrm{f2f}_j \left( X^q, X^{t^+} \right) \mathbf{x}_j^q \right], \quad j=1, \dots, N,
\end{aligned}
&&
\end{flalign}
\begin{flalign}
\begin{aligned}
    \frac{\partial \mathcal{L}}{\partial \mathbf{x}_0^{t_i^-}} &= p_i^- \frac{\partial \phi \left( X^q, X^{t_i^-} \right)}{\partial \mathbf{x}_0^{t_i^-}} \\
    &= \frac{1}{\tau} p_i^- \left[ w^\mathrm{g2g} \left( X^q, X^{t_i^-} \right) \mathbf{x}_0^q
    + \sum_{j=1}^N w^\mathrm{f2g}_j \left( X^q, X^{t_i^-} \right) \mathbf{x}_j^q \right],
\end{aligned}
&&
\end{flalign}
\begin{flalign}
\begin{aligned}
    \frac{\partial \mathcal{L}}{\partial \mathbf{x}_j^{t_i^-}} &= p_i^- \frac{\partial \phi \left( X^q, X^{t_i^-} \right)}{\partial \mathbf{x}_j^{t_i^-}} \\
    &= \frac{1}{\tau} p_i^- \left[ w^\mathrm{g2f}_j \left( X^q, X^{t_i^-} \right) \mathbf{x}_0^q
    + w^\mathrm{f2f}_j \left( X^q, X^{t_i^-} \right) \mathbf{x}_j^q \right], \quad j = 1, \dots, N,
\end{aligned}
&&
\end{flalign}
where the classification probabilities of the target samples $p^+$ and $p_i^-$ are defined by \cref{eq:cls-prob}. We utilize the following fact during the derivation:
\begin{equation}
    p^+ + \sum_{i=1}^{B-1} p_i^- = 1.
\end{equation}

\section{Detailed Results on MMEB}
\label{sec:mmeb-detailed}

\begin{table}[ht]
  \caption{Detailed results on the MMEB benchmark. The OOD datasets are highlighted with a yellow background.}
  \label{tab:mmeb-detailed}
  \begin{center}
    \resizebox{\textwidth}{!}{
    \begin{tabular}{lcccccccccccc}
      \toprule
      \rowcolor{gray!30} & \textbf{CLIP} & \textbf{OpenCLIP} & \textbf{SigLIP} & \textbf{BLIP2} & \textbf{MagicLens} & \textbf{E5-V} & \textbf{UniIR} & \textbf{VLM2Vec} & \textbf{UniME-V2} & \textbf{UME-R1} & \textbf{ReMatch} & \textbf{\model} \\
      \midrule
      \rowcolor{orange!30} \textbf{Classification (10 tasks)} & & & & & & & & & & & & \\
      ImageNet-1K & 55.8 & 63.5 & 45.4 & 10.3 & 48.0 & 9.6 & 58.3 & 74.5 & 78.8 & 80.4 & 78.7 & 83.2 \\
      N24News & 34.7 & 38.6 & 13.9 & 36.0 & 33.7 & 23.4 & 42.5 & 80.3 & 66.6 & 82.3 & 81.0 & 78.8 \\
      HatefulMemes & 51.1 & 51.7 & 47.2 & 49.6 & 49.0 & 49.7 & 56.4 & 67.9 & 65.3 & 79.0 & 67.8 & 89.8 \\
      VOC2007 & 50.7 & 52.4 & 64.3 & 52.1 & 51.6 & 49.9 & 66.2 & 91.5 & 92.0 & 90.8 & 89.5 & 91.6 \\
      SUN397 & 43.4 & 68.8 & 39.6 & 34.5 & 57.0 & 33.1 & 63.2 & 75.8 & 78.7 & 80.3 & 78.7 & 83.8 \\
      \rowcolor{yellow!15} Place365 & 28.5 & 37.8 & 20.0 & 21.5 & 31.5 & 8.6 & 36.5 & 44.0 & 42.9 & 46.8 & 45.4 & 49.6 \\
      \rowcolor{yellow!15} ImageNet-A & 25.5 & 14.2 & 42.6 & 3.2  & 8.0  & 2.0 & 9.8 & 43.6 & 48.0 & 53.9 & 48.4 & 59.1 \\
      \rowcolor{yellow!15} ImageNet-R & 75.6 & 83.0 & 75.0 & 39.7 & 70.9 & 30.8 & 66.2 & 79.8 & 89.3 & 90.1 & 83.5 & 90.1 \\
      \rowcolor{yellow!15} ObjectNet  & 43.4 & 51.4 & 40.3 & 20.6 & 31.6 & 7.5 & 32.2& 39.6 & 73.1 & 42.3 & 59.4 & 69.8 \\
      \rowcolor{yellow!15} Country-211 & 19.2 & 16.8 & 14.2 & 2.5  & 6.2  & 3.1 & 11.3 & 14.7 & 19.8 & 25.0 & 25.4 & 30.4 \\
      \textit{All Classification} & 42.8 & 47.8 & 40.3 & 27.0 & 38.8 & 21.8 & 44.3 & 61.2 & 65.3 & 67.1 & 65.8 & 72.6 \\
      \midrule
      \rowcolor{blue!30} \textbf{VQA (10 tasks)} & & & & & & & & & & & & \\
      OK-VQA & 7.5 & 11.5 & 2.4  & 8.7  & 12.7 & 8.9 & 25.4 & 69.0 & 71.9 & 71.7 & 72.6 & 75.4 \\
      A-OKVQA & 3.8 & 3.3 & 1.5 & 3.2 & 2.9 & 5.9 & 8.8 & 54.4 & 71.4 & 58.7 & 63.6 & 72.2 \\
      DocVQA & 4.0 & 5.3 & 4.2 & 2.6 & 3.0 & 1.7 & 6.2 & 52.0 & 92.6 & 93.8 & 95.8 & 95.4 \\
      InfographicsVQA & 4.6 & 4.6 & 2.7 & 2.0 & 5.9 & 2.3 & 4.6 & 30.7 & 63.5 & 79.2 & 82.9 & 73.4 \\
      ChartQA & 1.4 & 1.5 & 3.0 & 0.5 & 0.9 & 2.4 & 1.6 & 34.8 & 55.8 & 75.1 & 75.5 & 68.8 \\
      Visual7W & 4.0 & 2.6 & 1.2 & 1.3 & 2.5 & 5.8 & 14.5 & 49.8 & 62.5 & 55.2 & 66.9 & 64.4 \\
      \rowcolor{yellow!15} ScienceQA & 9.4 & 10.2 & 7.9  & 6.8  & 5.2 & 3.6 & 12.8 & 42.1 & 54.0 & 53.7 & 59.5 & 65.3 \\
      \rowcolor{yellow!15} VizWiz & 8.2 & 6.6 & 2.3 & 4.0 & 1.7 & 2.6 & 24.3 & 43.0 & 53.7 & 51.6 & 53.5 & 54.9 \\
      \rowcolor{yellow!15} GQA & 41.3 & 52.5 & 57.5 & 9.7  & 43.5 & 7.8 & 48.8 & 61.2 & 69.5 & 69.3 & 77.9 & 70.0 \\
      \rowcolor{yellow!15} TextVQA & 7.0 & 10.9 & 1.0 & 3.3 & 4.6 & 8.2 & 15.1 & 62.0 & 84.5 & 83.5 & 87.6 & 86.3 \\
      \textit{All VQA} & 9.1 & 10.9 & 8.4 & 4.2 & 8.3 & 4.9 & 16.2 & 49.9 & 67.6 & 69.2 & 73.6 & 72.6 \\
      \midrule
      \rowcolor{green!30} \textbf{Retrieval (12 tasks)} & & & & & & & & & & & & \\
      VisDial & 30.7 & 25.4 & 21.5 & 18.0 & 24.8 & 9.2 & 42.2 & 80.9 & 84.2 & 80.7 & 85.8 & 83.7 \\
      CIRR & 12.6 & 15.4 & 15.1 & 9.8  & 39.1 & 6.1 & 51.3 & 49.9 & 65.5 & 55.3 & 59.8 & 60.4 \\
      VisualNews\_t2i & 78.9 & 74.0 & 51.0 & 48.1 & 50.7 & 13.5 & 74.3 & 75.4 & 77.3 & 76.8 & 79.2 & 79.3 \\
      VisualNews\_i2t & 79.6 & 78.0 & 52.4 & 13.5 & 21.1 & 8.1 & 76.8 & 80.0 & 79.2 & 82.0 & 83.1 & 83.2 \\
      MSCOCO\_t2i & 59.5 & 63.6 & 58.3 & 53.7 & 54.1 & 20.7 & 68.5 & 75.7 & 79.1 & 78.3 & 81.2 & 81.6 \\
      MSCOCO\_i2t & 57.7 & 62.1 & 55.0 & 20.3 & 40.0 & 14.0 & 72.1 & 73.1 & 75.2 & 71.4 & 76.0 & 79.4 \\
      NIGHTS & 60.4 & 66.1 & 62.9 & 56.5 & 58.1  & 4.2 & 66.2 & 65.5 & 68.1 & 68.1 & 69.5 & 71.2 \\
      WebQA & 67.5 & 62.1 & 58.1 & 55.4 & 43.0 & 17.7 & 89.6 & 87.6 & 90.6 & 90.9 & 90.1 & 91.5 \\
      \rowcolor{yellow!15} FashionIQ & 11.4 & 13.8 & 20.1 & 9.3  & 11.2 & 2.8 & 40.2 & 16.2 & 26.4 & 23.4 & 29.6 & 23.7 \\
      \rowcolor{yellow!15} Wiki-SS-NQ & 55.0 & 44.6 & 55.1 & 28.7 & 18.7 & 8.6 & 12.2 & 60.2 & 71.2 & 72.5 & 69.5 & 67.7 \\
      \rowcolor{yellow!15} OVEN & 41.1 & 45.0 & 56.0 & 39.5 & 1.6  & 5.9 & 69.4 & 56.5 & 68.0 & 71.4 & 73.2 & 77.2 \\
      \rowcolor{yellow!15} EDIS & 81.0 & 77.5 & 23.6 & 54.4 & 62.6 & 26.8 & 79.2 & 87.8 & 88.2 & 92.0 & 92.5 & 86.4 \\
      \textit{All Retrieval} & 53.0 & 52.3 & 31.6 & 33.9 & 35.4 & 11.5 & 61.8 & 67.4 & 72.9 & 71.9 & 74.1 & 73.8 \\
      \midrule
      \rowcolor{purple!30} \textbf{Visual Grounding (4 tasks)} & & & & & & & & & & & & \\
      MSCOCO & 33.8 & 34.5 & 46.4 & 28.9 & 22.1 & 10.8 & 46.6 & 80.6 & 78.2 & 72.7 & 84.2 & 83.6 \\
      \rowcolor{yellow!15} RefCOCO & 56.9 & 54.2 & 70.8 & 47.4 & 22.8 & 11.9 & 67.8 & 88.7 & 94.6 & 91.4 & 95.8 & 92.9 \\
      \rowcolor{yellow!15} RefCOCO-matching & 61.3 & 68.3 & 50.8 & 59.5 & 35.6 & 38.9 & 62.9 & 84.0 & 91.4 & 91.1 & 94.0 & 94.0 \\
      \rowcolor{yellow!15} Visual7W-pointing & 55.1 & 56.3 & 70.1 & 52.0 & 23.4 & 14.3 & 71.3 & 90.9 & 93.8 & 84.2 & 96.0 & 87.7 \\
      \textit{All Visual Grounding} & 51.8 & 53.3 & 59.5 & 47.0 & 26.0 & 19.0 & 65.3 & 86.1 & 90.2 & 84.9 & 92.5 & 89.6 \\
      \midrule
      \rowcolor{cyan!15} \textbf{Final Score (36 tasks)} & & & & & & & & & & & & \\
      All & 37.8 & 39.7 & 34.8 & 25.2 & 27.8 & 13.3 & 44.7 & 62.9 & 71.2 & 71.3 & 73.7 & 74.9 \\
      All IND & 37.1 & 39.3 & 32.3 & 25.3 & 31.0 & 14.9 & 47.1 & 67.5 & 74.8 & 76.1 & 78.1 & 79.5 \\
      All OOD & 38.7 & 40.2 & 38.0 & 25.1 & 23.7 &  11.5 & 41.7 & 57.1 & 66.7 & 65.1 & 68.2 & 69.1 \\
      \bottomrule
    \end{tabular}
    }
  \end{center}
\end{table}

\section{Additional Experiments on the Impact of Hyperparameters}
\label{sec:hyper}

\begin{table}[ht]
  \caption{Hyperparameter impact analysis on the MMEB benchmark.}
  \label{tab:hyper-mmeb}
  \begin{center}
    \begin{tabular}{lccccccc}
      \toprule
      \multirow{2}{*}{\textbf{Model}} & \multicolumn{4}{c}{\textbf{Per Meta-Task Score}} & \multicolumn{3}{c}{\textbf{Average Score}} \\
      \cmidrule(lr){2-5} \cmidrule(lr){6-8}
      & \textbf{Classification} & \textbf{VQA} & \textbf{Retrieval} & \textbf{Grounding} & \textbf{IND} & \textbf{OOD} & \textbf{Overall} \\
      \midrule
      \textbf{\# of datasets} & 10 & 10 & 12 & 4 & 20 & 16 & 36 \\
      \midrule
      $M=N=10$, $\alpha=20$ (default) & \textbf{72.6} & \textbf{72.6} & 73.8 & \textbf{90.0} & \textbf{79.5} & \textbf{69.1} & \textbf{74.9} \\
      \midrule
      \multicolumn{8}{c}{\textit{$M, N$ Modification}} \\
      \midrule
      $M=N=5$, $\alpha=20$ & 72.1 & 71.7 & 72.8 & 88.9 & 78.6 & 68.4 & 74.1 \\
      $M=N=3$, $\alpha=20$ & 71.8 & 71.7 & 72.9 & 89.2 & 78.3 & 68.8 & 74.1 \\
      $M=N=0$, $\alpha=20$ (QQMM) & 69.9 & 70.0 & 72.1 & 86.0 & 77.2 & 66.6 & 72.5 \\
      \midrule
      \multicolumn{8}{c}{\textit{$\alpha$ Modification}} \\
      \midrule
      $M=N=10$, $\alpha=10$ & 72.1 & 72.3 & \textbf{73.9} & 89.5 & 79.1 & \textbf{69.1} & 74.7 \\
      $M=N=10$, $\alpha=5$ & 72.3 & 72.4 & 73.5 & 89.1 & 79.1 & 68.9 & 74.6 \\
      $M=N=10$, $\alpha=0$ (w/o EGA) & 72.5 & 72.2 & 73.4 & 89.3 & 79.0 & \textbf{69.1} & 74.6 \\
      \bottomrule
    \end{tabular}
  \end{center}
\end{table}

\begin{table}[ht]
  \caption{Hyperparameter impact analysis on the MMVP-VLM benchmark.}
  \label{tab:hyper-mmvp}
  \begin{center}
    \begin{tabular}{lcccccccccc}
      \toprule
      \textbf{Model} & \textbf{Ori.} & \textbf{Pre.} & \textbf{Sta.} & \textbf{Qua.} & \textbf{Spa.} & \textbf{Col.} & \textbf{Str.} & \textbf{Tex.} & \textbf{Cam.} & \textbf{Average} \\
      \midrule
      $M=N=10$, $\alpha=20$ (default) & 60.0 & 43.3 & \textbf{73.3} & \textbf{73.3} & 46.7 & 83.3 & \textbf{66.7} & \textbf{50.0} & 56.7 & \textbf{61.5} \\
      \midrule
      \multicolumn{11}{c}{\textit{$M, N$ Modification}} \\
      \midrule
      $M=N=5$, $\alpha=20$ & 56.7 & \textbf{50.0} & 66.7 & 63.3 & 50.0 & 80.0 & 63.3 & 43.3 & 60.0 & 59.3 \\
      $M=N=3$, $\alpha=20$ & 56.7 & 43.3 & 70.0 & 70.0 & 50.0 & 76.7 & 60.0 & 43.3 & 56.7 & 58.5 \\
      $M=N=0$, $\alpha=20$ (QQMM) & 53.3 & 43.3 & \textbf{73.3} & 46.7 & 36.7 & 83.3 & 50.0 & 43.3 & 56.7 & 54.1 \\      
      \midrule
      \multicolumn{11}{c}{\textit{$\alpha$ Modification}} \\
      \midrule
      $M=N=10$, $\alpha=10$ & 56.7 & 46.7 & 70.0 & 70.0 & 46.7 & \textbf{86.7} & 63.3 & \textbf{50.0} & \textbf{63.3} & \textbf{61.5} \\
      $M=N=10$, $\alpha=5$ & 56.7 & 43.3 & 66.7 & 66.7 & 50.0 & 83.3 & 63.3 & 46.7 & \textbf{63.3} & 60.0 \\
      $M=N=10$, $\alpha=0$ (w/o EGA) & \textbf{63.3} & \textbf{50.0} & 70.0 & 63.3 & \textbf{53.3} & 80.0 & 63.3 & 36.7 & 60.0 & 60.0 \\
      \bottomrule
    \end{tabular}
  \end{center}
\end{table}

Additionally, we analyze the impact of key hyperparameters in \model, namely the number of fine-grained embedding modules $N$, the number of prompt tokens $M$, and the amplification magnitude $\alpha$. The main experiments in \cref{sec:mmeb,sec:mmvp} adopt the default settings of $M=N=10$ and $\alpha=20$. Subsequently, we adjust $M=N=\{3,5\}$ and $\alpha=\{5,10\}$ separately. The evaluation results on MMEB are presented in \cref{tab:hyper-mmeb}, and those on MMVP-VLM are shown in \cref{tab:hyper-mmvp}. Whether in general or fine-grained scenarios, \model~equipped with fine-grained embedding modules of varying scales $M,N$ achieves performance improvements of varying degrees compared to the pre-finetuned QQMM, which demonstrates the strong robustness of the perceptual fusion mechanism. Furthermore, due to the relatively low amplification magnitude, \model~with $\alpha=5$ shows no gain compared to the version without EGA. However, when $\alpha$ is increased to $10$, the performance improvement becomes evident.

\section{Time Cost Analysis}
\label{sec:time}

\begin{table}[ht]
    \caption{Training and inference time on the MMEB benchmark.}
  \label{tab:time}
  \begin{center}
    \begin{tabular}{lcc}
    \toprule
      \textbf{Model} & \textbf{Training Time per Sample (s)} & \textbf{Inference Time per Sample (s)} \\
      \midrule
      QQMM ($M=N=0$) & 2.41 & 0.658 \\
      \model~($M=N=3$) & 2.58 & 0.693 \\
      \model~($M=N=5$) & 2.70 & 0.716 \\
      \model~($M=N=10$) & 3.08 & 0.783 \\
      \bottomrule
    \end{tabular}
  \end{center}
\end{table}

During training and evaluation on MMEB, we record the total time cost $T_\mathrm{total}$ and calculate the per-sample training and inference time:
\begin{equation}
    T_\mathrm{sample} = \frac{T_\mathrm{total}}{N \cdot B},
\end{equation}
where $N$ is the number of batches, and $B$ is the batch size. In \cref{tab:time}, we report the per-sample time cost of \model~with different numbers of fine-grained embedding modules $N$ and different numbers of prompt tokens $M$, as well as that of the original QQMM. We observe that although increasing $M,N$ introduces additional time overhead, it is not substantial. Even in the case of $M=N=10$, it only increases the training time by approximately $28\%$ and the inference time by about $19\%$ compared to the original QQMM. In fact, for \model, the additional cost introduced by the fine-grained embedding modules and similarity aggregation is marginal compared to image processing.

\end{document}